\documentclass{bmvc2k-widemargin}

\title{Joint Calibration for Semantic Segmentation}

\addauthor{Holger Caesar}{holger.caesar@ed.ac.uk}{1}
\addauthor{Jasper Uijlings}{jrr.uijlings@ed.ac.uk}{1}
\addauthor{Vittorio Ferrari}{vittorio.ferrari@ed.ac.uk}{1}

\addinstitution{
 School of Informatics\\
 University of Edinburgh\\
 Edinburgh, UK
}

\runninghead{Caesar et al.}{Joint Calibration for Semantic Segmentation}

\newcommand{\mypar}[1]{\vspace{-4mm}\paragraph{#1}}
\DeclareMathOperator*{\argmax}{arg\,max}

\begin{document}

\maketitle

\begin{abstract}
Semantic segmentation is the task of assigning a class-label to each pixel in an image.
We propose a region-based semantic segmentation framework which handles both full and weak supervision, and addresses three common problems:
(1) Objects occur at multiple scales and therefore we should use regions at multiple scales.
    However, these regions are overlapping which creates conflicting class predictions at the pixel-level.
(2) Class frequencies are highly imbalanced in realistic datasets.
(3) Each pixel can only be assigned to a single class, which creates competition between classes.
We address all three problems with a joint calibration method which optimizes a multi-class loss defined over the final pixel-level output labeling, as opposed to simply region classification.
Our method outperforms the state-of-the-art on the popular SIFT Flow~\cite{liu11pami} dataset in both the fully and weakly supervised setting by a considerably margin (+6\% and +10\%, respectively).
\end{abstract}


\section{Introduction}
\label{sec:intro}

\begin{figure}[tb]
\centering
\vspace{-1mm}
\includegraphics[width=1.0\textwidth]{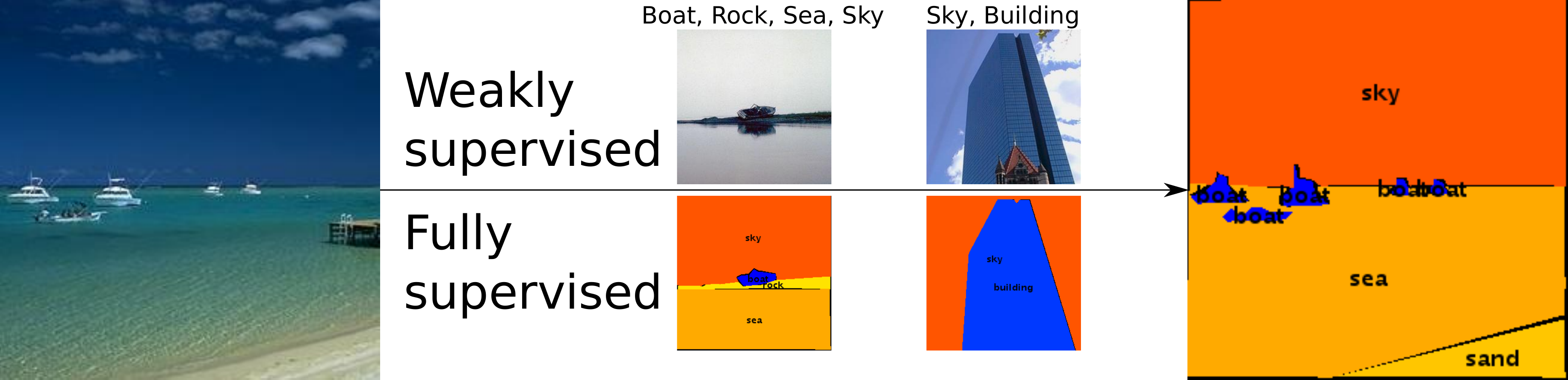}
\vspace{-6mm}
\caption{Semantic segmentation is the task of assigning class labels to all pixels in the image.
    During training, with full supervision we have ground-truth labels of all pixels. With weak
supervision we only have labels at the image-level.}
\vspace{-5mm}
\label{fig:task}
\end{figure}

Semantic segmentation is the task of assigning a class label to each pixel in an image (Fig.~\ref{fig:task}).
In the fully supervised setting, we have ground-truth labels for all pixels in the training images.
In the weakly supervised setting, class-labels are only given at the image-level. 
We tackle both settings in a single framework which builds on region-based classification. 

Our framework addresses three important problems common to region-based semantic
segmentation. First of all, objects naturally occur at different scales within an image~\cite{carreira10cvpr,uijlings13ijcv}.
Performing recognition at a single scale inevitably leads to regions covering only parts of an object which may have ambiguous appearance, such as \emph{wheels} or \emph{fur}, and to regions straddling over multiple objects, whose classification is harder due to their mixed appearance.
Therefore many recent methods operate on pools of regions computed at multiple scales, which have a much better chance of containing some regions covering complete objects~\cite{plath09icml,carreira10cvpr,carreira12eccv,li13cvpr,hariharan14eccv,girshick14cvpr,zhang15cvpr}.
However, this leads to overlapping regions which may lead to conflicting class predictions at the pixel-level. These conflicts need to be properly resolved. 

Secondly, classes are often unbalanced~\cite{farabet13pami,tighe13cvpr,vezhnevets11iccv,kekec14bmvc,xu14cvpr,yang14cvpr,long15cvpr,sharma14nips,sharma15cvpr,xu15cvpr,mostajabi15cvpr,byeon15cvpr,shuai15cvpr}:
``cars'' and ``grass'' are frequently found in images while ``tricycles'' and ``gravel'' are much rarer.
Due to the nature of most classifiers, without careful consideration these rare classes are largely ignored:
even if the class occurs in an image the system will rarely predict it.
Since class-frequencies typically follow a power-law distribution, this problem becomes increasingly important with the modern trend towards larger datasets with more and more classes.

Finally, classes compete: a pixel can only be assigned to a single class (e.g. it can not belong to both ``sky'' and ``airplane'').
To properly resolve such competition, a semantic segmentation framework should take into account predictions for multiple classes jointly. 

In this paper we address these three problems with a joint calibration method over an ensemble of SVMs, where the calibration parameters are optimized over all classes, and for the final evaluation criterion, i.e. the accuracy of pixel-level labeling, as opposed to simply region classification.
While each SVM is trained for a single class, their joint calibration deals with the competition between classes.
Furthermore, the criterion we optimize for explicitly accounts for class imbalance.
Finally, competition between overlapping regions is resolved through maximization: each pixel is assigned the
highest scoring class over all regions covering it. We jointly calibrate the SVMs for optimal pixel labeling \emph{after} this maximization, which effectively takes into account conflict resolution between overlapping regions.
Experiments on the popular SIFT Flow~\cite{liu11pami} dataset show a considerable improvement over the state-of-the-art in both the fully and weakly supervised setting (+6\% and +10\%, respectively).

\vspace{-4mm}

\section{Related work}

Early works on semantic segmentation used pixel- or patch-based features over which they define a Condition Random Field (CRF)~\cite{shotton09ijcv,verbeek07cvpr}.
Many modern successful works use region-level representations, both in the fully supervised~\cite{boix12ijcv,carreira12eccv,hariharan14eccv,girshick14cvpr,li13cvpr,plath09icml,tighe10eccv,tighe11iccv,tighe14cvpr,yang14cvpr,sharma14nips,sharma15cvpr,george15cvpr,mostajabi15cvpr}
and weakly supervised~\cite{vezhnevets11iccv,vezhnevets12cvpr,xu14cvpr,zhang14tm,xu15cvpr,zhang15cvpr} settings.
A few recent works use CNNs to learn a direct mapping from image to pixel labels~\cite{farabet13pami,sharma14nips,sharma15cvpr,pinheiro14icml,pinheiro15cvpr,shuai15cvpr,zheng15arxiv,papandreou15arxiv,schwing15arxiv,lin15arxiv}, 
although some of them~\cite{farabet13pami,sharma14nips,sharma15cvpr} use region-based post-processing to impose label smoothing and to better respect object boundaries.
Other recent works use CRFs to refine the CNN pixel-level predictions~\cite{zheng15arxiv,schwing15arxiv,lin15arxiv,chen15iclr,papandreou15arxiv}. 
In this work we focus on region-based semantic segmentation, which we discuss in light of the three problems raised in the introduction.

\mypar{Overlapping regions.}
Traditionally, semantic segmentation systems use superpixels~\cite{boix12ijcv,tighe10eccv,tighe11iccv,tighe14cvpr,yang14cvpr,george15cvpr,mostajabi15cvpr,sharma14nips,sharma15cvpr}, 
which are non-overlapping regions resulting from a single-scale oversegmentation.
However, appearance-based recognition of superpixels is difficult as they typically capture only parts of objects, rather than complete objects.
Therefore, many recent methods use overlapping multi-scale regions~\cite{carreira10cvpr,carreira12eccv,hariharan14eccv,girshick14cvpr,li13cvpr,plath09icml,zhang15cvpr}.
However, these may lead to conflicting class predictions at the pixel-level.
Carreira et al.~\cite{carreira12eccv} address this simply by taking the maximum score over all regions containing a pixel.
Both Hariharan et al.~\cite{hariharan14eccv} and Girshick et al.~\cite{girshick14cvpr} use non-maximum suppression, which may give problems for nearby or interacting objects~\cite{li13cvpr}.
Li et al.~\cite{li13cvpr} predict class overlap scores for each region at each scale.
Then they create superpixels by intersecting all regions.
Finally, they assign overlap scores to these superpixels using maximum composite likelihood (i.e.  taking all multi-scale predictions into account).
Plath et al.~\cite{plath09icml} use classification predictions over a segmentation hierarchy to induce label consistency between parent and child regions within a tree-based CRF framework.
After solving their CRF formulation, only the smallest regions (i.e.  leaf-nodes) are used for class prediction.
In the weakly supervised setting, most works use superpixels~\cite{vezhnevets11iccv,vezhnevets12cvpr,xu14cvpr,xu15cvpr} and so do not encounter problems of conflicting predictions.
Zhang et al.~\cite{zhang14tm} use overlapping regions to enforce a form of class-label smoothing, but they all have the same scale.
A different Zhang et al.~\cite{zhang15cvpr} use overlapping region proposals at multiple scales in a CRF.

\mypar{Class imbalance.}
As the PASCAL VOC dataset~\cite{everingham10ijcv} is relatively balanced, most works that experiment on it did not explicitly address this issue~\cite{boix12ijcv,carreira12eccv,hariharan14eccv,girshick14cvpr,li13cvpr,plath09icml,long15cvpr,pinheiro15cvpr,zheng15arxiv,schwing15arxiv,lin15arxiv,chen15iclr}. 
On highly imbalanced datasets such as SIFT Flow~\cite{liu11pami}, Barcelona~\cite{tighe10eccv} and LM+SUN~\cite{tighe13cvpr}, rare classes pose a challenge.
This is observed and addressed by Tighe et al.~\cite{tighe13cvpr} and Yang et al.~\cite{yang14cvpr}: 
for a test image, only a few training images with similar context are used to provide class predictions,
but for rare classes this constraint is relaxed and more training images are used.
Vezhnevets et al.~\cite{vezhnevets11iccv} balance rare classes by normalizing scores for each class to range $[0,1]$.
A few works~\cite{mostajabi15cvpr,xu14cvpr,xu15cvpr} balance classes by using an inverse class frequency weighted loss function.

\mypar{Competing classes.}
Several works train one-vs-all classifiers separately and resolve labeling through maximization~\cite{carreira12eccv,hariharan14eccv,girshick14cvpr,li13cvpr,plath09icml,tighe13cvpr,mostajabi15cvpr,pinheiro15cvpr}. 
This is suboptimal since the scores of different classes may not be properly calibrated. 
Instead, Tighe et al.~\cite{tighe10eccv,tighe13cvpr} and Yang et al.~\cite{yang14cvpr} use Nearest Neighbor classification which is inherently multi-class.
In the weakly supervised setting appearance models are typically trained in isolation and remain uncalibrated~\cite{vezhnevets11iccv,vezhnevets11iccv,zhang14tm,xu14cvpr,xu15cvpr}. 
To the best of our knowledge, Boix et al.~\cite{boix12ijcv} is the only work in semantic segmentation to perform joint calibration of SVMs. While this enables to handle competing classes, in their work they use non-overlapping regions.
In contrast, in our work we use overlapping regions where conflicting predictions are resolved through maximization.
In this setting, joint calibration is particularly important, as we will show in Sec.~\ref{sec:experiments}.
As another difference, Boix et al.~\cite{boix12ijcv} address only full supervision whereas we address both full and weak supervision in a unified framework.

\vspace{-3mm}
\section{Method}
\label{sec:method}
\subsection{Model}
\label{sec:model}

We represent an image by a set of overlapping regions~\cite{uijlings13ijcv} described by CNN features~\cite{girshick14cvpr} (Sec.~\ref{sec:implementation-details}).
Our semantic segmentation model infers the label $o_p$ of each pixel $p$ in an image:
\begin{align}
 o_p = \argmax_{c , \; r \, \ni p} \; \sigma \! \left( w_c \cdot x_r, \; a_c, b_c \right)
 \label{eq:model}
\end{align}
As appearance models, we have a separate linear SVM $w_c$ per class $c$. These SVMs score the features $x_r$ of each region $r$.
The scores are calibrated by a sigmoid function $\sigma$, with different parameters $a_c, b_c$ for each class $c$.
The $\argmax$ returns the class $c$ with the highest score over all regions that contain pixel $p$.
This involves maximizing over classes for a region, and over the regions that contain $p$.

During training we find the SVM parameters $w_c$ (Sec.~\ref{sec:svm_training}) and calibration parameters $a_c$ and $b_c$ (Sec.~\ref{sec:jc}).
The training of the calibration parameters takes into account the effects of the two maximization
operations, as they are optimized for the output pixel-level labeling performance (as opposed to
simply accuracy in terms of region classification).

\subsection{SVM training}
\label{sec:svm_training}
\vspace{+3mm}

\mypar{Fully supervised.}
In this setting we are given ground-truth pixel-level labels for all images in the training set
(Fig.~\ref{fig:task}). This leads to a natural subdivision into ground-truth regions, i.e. non-overlapping
regions perfectly covering a single class. We use these as positive training samples. However, such
idealized samples are rarely encountered at test time since there we have only imperfect region
proposals~\cite{uijlings13ijcv}.  Therefore we use as additional positive samples for a class all region proposals
which overlap heavily with a ground-truth region of that class~(i.e. Intersection-over-Union greater than 50\%~\cite{everingham10ijcv}).
As negative samples, we use all regions from all images that do not contain that class. 
In the SVM loss function we apply inverse frequency weighting in terms of the number of positive
and negative samples.

\mypar{Weakly supervised.}
In this setting we are only given image-level labels on the training images (Fig.~\ref{fig:task}).
Hence, we treat region-level labels as latent variables which are updated using an alternated 
optimization process (as in~\cite{vezhnevets11iccv,vezhnevets12cvpr,xu14cvpr,zhang15cvpr,xu15cvpr}).
To initialize the process, we use as positive samples for a class all regions in all images containing it.
At each iteration we alternate between training SVMs based on the current region labeling and updating the labeling
based on the current SVMs (by assigning to each region the label of the highest scoring class).
In this process we keep our negative samples constant, i.e. all regions from all images that do not contain the target class.
In the SVM loss function we apply inverse frequency weighting in terms of the number of positive
and negative samples.

\vspace{-2mm}
\subsection{Joint Calibration}
\label{sec:jc}

We now introduce our joint calibration procedure, which addresses three common
problems in semantic segmentation: (1) conflicting predictions of overlapping regions, (2) class
imbalance, and (3) competition between classes.

\begin{figure*}
\begin{center}
\includegraphics[width=1.0\textwidth]{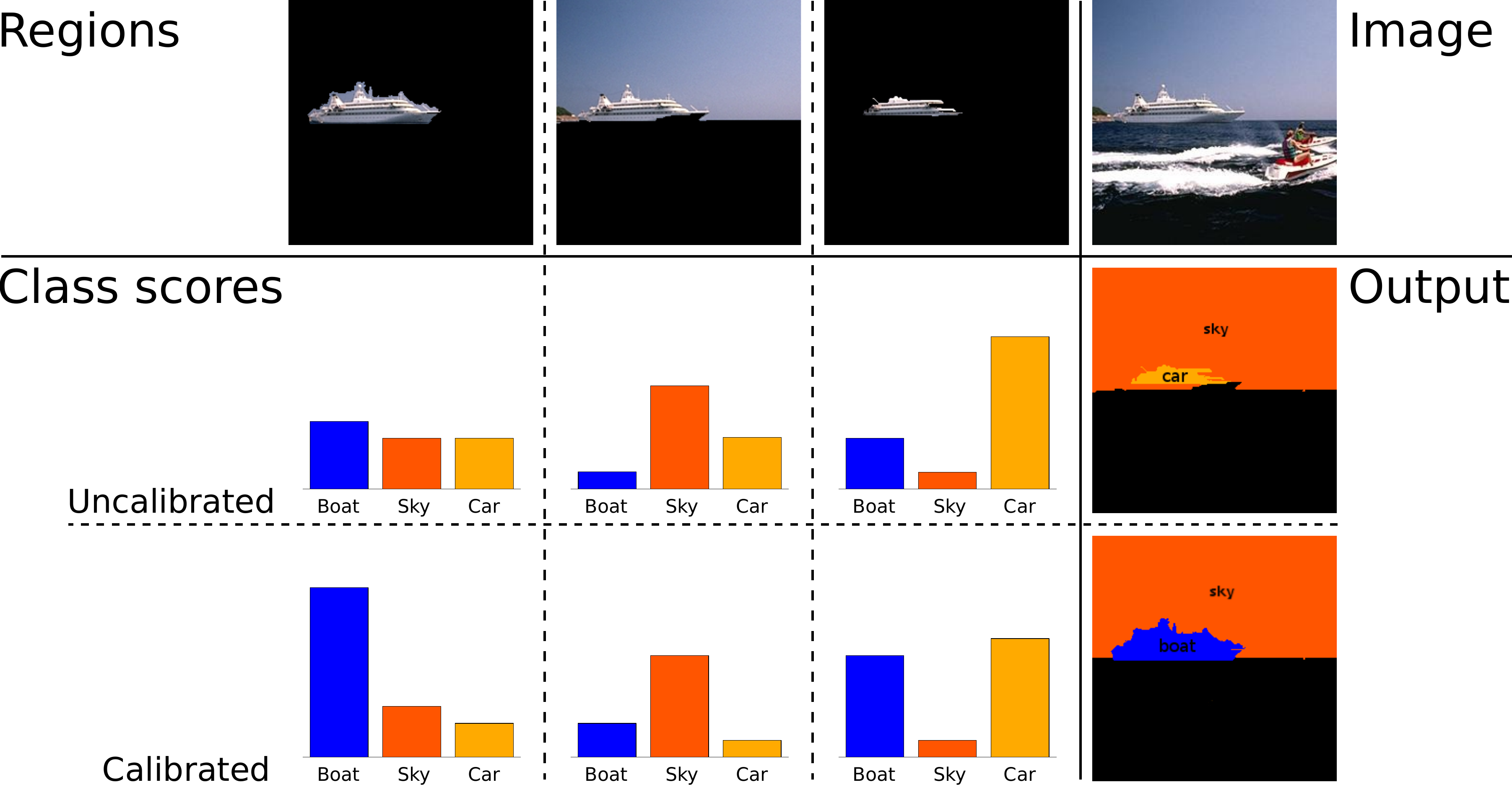}

\end{center}
\vspace{-4mm}
\caption{
The first row shows multiple region proposals (left) extracted from an image (right).
The following rows show the per-class SVM scores of each region (left) and the pixel-level labeling (right).
Row 2 shows the results before and row 3 after joint calibration.
}
\vspace{-3mm}
\label{fig:jointcalibration}
\end{figure*}

To better understand the problem caused by overlapping regions, consider the example of
Fig.~\ref{fig:jointcalibration}. It shows three overlapping regions, each with different class
predictions. The final goal of semantic segmentation is to output a pixel-level labeling, which is
evaluated in terms of pixel-level accuracy. In our framework we employ a winner-takes all principle: each
pixel takes the class of the highest scored region which contains it.  Now, using uncalibrated
SVMs is problematic (second row in Fig.~\ref{fig:jointcalibration}).
SVMs are trained to predict class labels at the region-level, not the pixel-level. 
However, different regions have different area, and, most importantly, not all regions contribute all of their area to the final pixel-level labeling:
Predictions of small regions may be completely suppressed by bigger regions (e.g. in Fig.~\ref{fig:jointcalibration}, row 3, the inner-boat region is suppressed by the prediction of the complete boat).
In other cases, bigger regions may be \emph{partially} overwritten by smaller regions (e.g. in Fig.~\ref{fig:jointcalibration} the boat region partially overwrites the prediction of the larger boat+sky region).
Furthermore, the SVMs are trained in a one-vs-all manner and are unaware of other classes.
Hence they are unlikely to properly resolve competition between classes even within a single region.
The problems above show that without calibration, the SVMs are optimized for the wrong criterion.
We propose to jointly calibrate SVMs for the correct criterion, which corresponds better to the evaluation measure typically used for semantic segmentation (i.e. pixel labeling accuracy averaged over classes).
We do this by applying sigmoid functions $\sigma$ to all SVM outputs:
\begin{equation}
  \sigma \! \left( w_c \cdot x_r, \; a_c, b_c \right) = (1+\exp{\left(a_c \cdot w_c
    \cdot x_r + b_c \right)})^{-1} \, 
\label{eq:sigmoid}
\end{equation}
where $a_c,b_c$ are the calibration parameters for class $c$. We calibrate the parameters of all classes jointly by minimizing a loss function $\mathcal{L}(o,l)$, where $o$ is the pixel labeling output of our method on the full training
set~($o = \{o_p; \, p = 1 \ldots P \}$) and $l$ the ground-truth labeling.

We emphasize that the pixel labeling output $o$ is the result {\em after} the maximization over
classes and regions in Eq.~\eqref{eq:model}. Since we optimize for the accuracy of this final output labeling, and we do so jointly over classes, our calibration procedure takes into account both problems of conflicting class predictions between
overlapping regions and competition between classes. Moreover, we also address the problem of class imbalance, as we compensate for it in our loss functions below.

\mypar{Fully supervised loss.}
In this setting our loss directly evaluates the desired performance measure, which is typically
pixel labeling accuracy averaged over classes~\cite{tighe10eccv,sharma14nips,yang14cvpr,farabet13pami,long15cvpr}
\begin{equation}
 \mathcal{L} (o, l) = 1 - \frac{1}{C} \sum_{c = 1}^{C} \frac{1}{P_c} \sum_{p ; \, l_p = c} [l_{p} = o_{p}]
\end{equation}
where $l_p$ is the ground-truth label of pixel $p$, $o_p$ is the output pixel label, $P_c$ is the
number of pixels with ground-truth label $c$, and $C$ is the number of classes. $[ \cdot ]$ is $1$
if the condition is true and $0$ otherwise. The inverse frequency weighting factor $1/P_c$ deals
with class imbalance.

\mypar{Weakly supervised loss.}
Also in this setting the performance measure is typically
class-average pixel accuracy~\cite{vezhnevets11iccv,vezhnevets12cvpr,xu15cvpr,zhang15cvpr}.  Since we
do not have ground-truth pixel labels, we cannot directly evaluate it.  We do however have a set of
ground-truth image labels $l_i$ which we can compare against.  We first aggregate the output pixel
labels $o_p$ over each image $m_i$ into output image labels $o_i = \cup_{p \in m_i} \, o_p$. Then we
define as loss the difference between the ground-truth label set $l_i$ and the output label set $o_i$,
measured by the Hamming distance between their binary vector representations
\begin{equation}
  \vspace{-2mm}
  \mathcal{L} \left( o, l \right) = \sum_{i = 1}^{I} \sum_{c = 1}^{C} \frac{1}{I_c}
    \left| l_{i, \,c} - o_{i, \, c} \right| 
  \label{eq:ws_loss}
\end{equation}
where $l_{i,c}=1$ if label $c$ is in $l_i$, and $0$ otherwise (analog for $o_{i,c}$). $I$ is the
total number of training images. $I_c$ is the
number of images having ground-truth label $c$, so the loss is weighted by the inverse frequency of
class labels, measured at the image-level.  Note how also in this setting the loss looks at
performance after the maximization over classes and regions (Eq.~\eqref{eq:model}).

\begin{figure*}
\begin{center}
\includegraphics[width=1.0\textwidth]{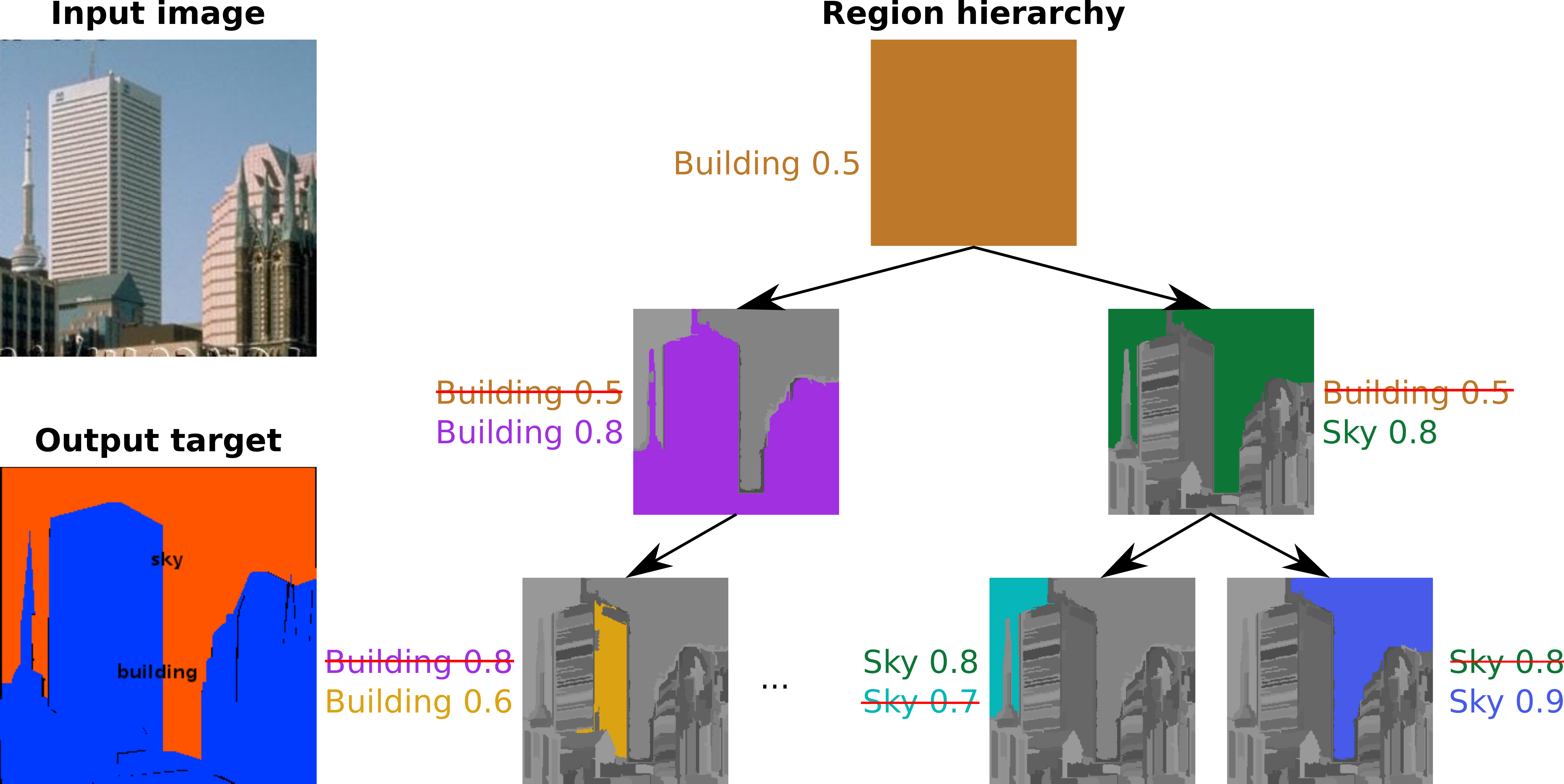}
\caption{
Our efficient evaluation algorithm uses the bottom-up structure of Selective Search region proposals to simplify the spatial maximization.
We start from the root node and propagate the maximum score with its corresponding label down the tree.
We label the image based on the labels of its superpixels~(leaf nodes).
}
\label{fig:efficient_evaluation}
\end{center}
\vspace{-6mm}
\end{figure*}

\mypar{Optimization.}
We want to minimize our loss functions over the calibration parameters $a_c, b_c$ of all classes.
This is hard, because the output pixel labels $o_p$ depend on these parameters in a complex manner due to the max over classes and regions in Eq.~\eqref{eq:model}, and because of the set-union aggregation in the case of the weakly supervised loss.
Therefore, we apply an approximate minimization algorithm based on coordinate descent. Coordinate
descent is different from gradient descent in that it can be used on arbitrary loss functions that
are not differentiable, as it only requires their evaluation for a given setting of parameters.

Coordinate descent iteratively applies line search to optimize the loss over a single parameter at a time, keeping all others fixed. This process cycles through all parameters until convergence. 
As initialization we use constant values~($a_c = -7, \; b_c = 0$).
During line search we consider 10 equally spaced values~($a_c$ in $[-12, -2], \; b_c$ in $[-10, 10]$).

This procedure is guaranteed to converge to a local minimum on the search grid. While this might not be the global optimum, in repeated trials we found the results to be rather insensitive to initialization. Furthermore, in our experiments the number of iterations was roughly proportional to the number of parameters.

\mypar{Efficient evaluation.}
\label{par:efficient_evaluation}
On a typical training set with $C=30$ classes, our joint calibration procedure evaluates the loss thousands of times.
Hence, it is important to evaluate pixel-level accuracy quickly.
As the model involves a maximum over classes and a maximum over regions at every pixel, a naive per-pixel implementation would be prohibitively expensive.
Instead, we propose an efficient technique that exploits the nature of the Selective Search region
proposals~\cite{uijlings13ijcv}, which form a bottom-up hierarchy starting from superpixels.
As shown in Fig.~\ref{fig:efficient_evaluation}, we start from the region proposal that contains the entire image~(root node).
Then we propagate the maximum score over all classes down the region hierarchy. 
Eventually we assign to each superpixel~(leaf nodes) the label with the highest score over all regions that contain it.
This label is assigned to all pixels in the superpixel.
To compute class-average pixel accuracy, we normally need to compare each pixel label to the ground-truth label.
However since we assign the same label to all pixels in a superpixel, we can precompute the ground-truth label distribution for each superpixel and use it as a lookup table.
This reduces the runtime complexity for an image from $O(P_i \cdot R_i \cdot C)$ to $O(R_i \cdot
C)$, where $P_i$ and $R_i$ are the number of pixels and regions in an image respectively, and $C$ is the number of classes. 

\mypar{Why no Platt scaling.}
At this point the reader may wonder why we do not simply use Platt
scaling~\cite{platt1999} as is commonly done in many applications. Platt scaling is
used to convert SVM scores to range $[0,1]$ using sigmoid functions, as in
Eq.~\eqref{eq:sigmoid}. 
However, in Platt scaling the parameters $a_c,b_c$ are optimized for each class
in isolation, ignoring class competition.
The loss function $\mathcal{L}_c$ in Platt scaling is the cross-entropy function
\begin{equation}
\mathcal{L}_c \left( \sigma_c, \, l \right) = - \sum_r t_{r, \,c} \log(\sigma_c(x_r))
    + (1 - t_{r, \,c}) \log(1 - \sigma_c(x_r)) \; 
\vspace{-2mm}
\end{equation}
where $N_+$ is the number of positive samples, $N_-$ the number of negative samples,
and
$t_{r, \,c} = \frac{N_+ + 1}{N_+ + 2}$ if $l_r = c$ or $t_{r, \,c} = \frac{1}{N_- + 2}$ otherwise;
$l_r$ is the region-level label. 
This loss function is inappropriate for semantic segmentation because it is
defined in terms of accuracy of training samples, which are regions, rather than in terms of the
final pixel-level accuracy.
Hence it ignores the problem of overlapping regions. There is also no inverse frequency term to deal with
class imbalance. We experimentally compare our method with Platt scaling in Sec.~\ref{sec:experiments}.

\subsection{Implementation Details}
\label{sec:implementation-details}
\vspace{+4mm}
\mypar{Region proposals.}
We use Selective Search~\cite{uijlings13ijcv} region proposals using a subset of the ``Fast'' mode:
we keep the similarity measures, but we restrict the scale parameter $k$ to 100 and the color-space to RGB. This leads to two bottom-up hierarchies of one initial oversegmentation~\cite{felzenszwalb04ijcv}.  

\mypar{Features.}
\label{par:features}
We show experiments with features generated by two CNNs (AlexNet~\cite{krizhevsky12nips}, VGG16~\cite{simonyan15iclr}) using the Caffe implementations \cite{jia13caffe}. We use the R-CNN~\cite{girshick14cvpr} framework for AlexNet, and Fast R-CNN~(FRCN)~\cite{girshick15arxiv} for VGG16, in order to maintain high computational efficiency.
Regions are described using all pixels in a tight bounding box.
Since regions are free-form, Girshick et al.~\cite{girshick14cvpr} additionally propose to set pixels not belonging to the region to zero (i.e. not affecting the convolution). However, in our experiments this did not improve results so we do not use it. 
For the weakly supervised setting we use the CNNs pre-trained for image classification on ILSVRC 2012~\cite{russakovsky15ijcv}. 
For the fully supervised setting we finetune them on the training set of SIFT Flow~\cite{liu11pami} (i.e. the semantic segmentation dataset we experiment on).
For both settings, following~\cite{girshick14cvpr} we use the output of the fc6 layer of the CNN as features.

\mypar{SVM training.}
\label{sec:classifier_training}
Like~\cite{girshick14cvpr} we set the regularization parameter C to a fixed value in all our experiments.
The SVMs minimize the L2-loss for region classification.
We use hard-negative mining to reduce memory consumption.

\section{Experiments}
\vspace{+3mm}
\label{sec:experiments}

\mypar{Datasets.}
We evaluate our method on the challenging SIFT Flow dataset~\cite{liu11pami}.
It consists of 2488 training and 200 test images, pixel-wise annotated with 33 class labels.
The class distribution is highly imbalanced in terms of overall region count as well as pixel count.
As evaluation measure we use the popular class-average pixel accuracy~\cite{tighe10eccv,tighe14cvpr,yang14cvpr,farabet13pami,long15cvpr,sharma15cvpr,pinheiro14icml,vezhnevets12cvpr,xu15cvpr,zhang15cvpr}.
For both supervision settings we report results on the test set.

\begin{table}[t]
\resizebox{0.5\textwidth}{!}{
\centering
\begin{tabular}{ | l | r | }
\hline
Method						&	Class Acc.	\\ 
\hline
\hline
Byeon et al.~\cite{byeon15cvpr}			&	22.6\%		\\ 
Tighe et al.~\cite{tighe10eccv}			&	29.1\%		\\ 
Pinheiro et al.~\cite{pinheiro14icml}		&	30.0\%		\\ 
Shuai et al.~\cite{shuai15cvpr}			&	39.7\%		\\
Tighe et al.~\cite{tighe13cvpr}			&	41.1\%		\\ 
Keke\c{c} et al.~\cite{kekec14bmvc}		&	45.8\%		\\
Sharma et al.~\cite{sharma14nips}		&	48.0\%		\\ 
Yang et al.~\cite{yang14cvpr}			&	48.7\%		\\ 
George et al.~\cite{george15cvpr}		&	50.1\%		\\ 
Farabet et al.~\cite{farabet13pami}		& 	50.8\%		\\ 
Long et al.~\cite{long15cvpr}			&	51.7\%		\\ 
Sharma et al.~\cite{sharma15cvpr}		&	52.8\%		\\ 
\hline 
Ours SVM \;\;\;\;\;\;(AlexNet)			& 	28.7\%		\\ 
Ours SVM+PS  (AlexNet)				& 	27.7\%		\\ 
Ours SVM+JC (AlexNet)				& 	55.6\%		\\ 
Ours SVM+JC (VGG16)				& \textbf{59.2\%}	\\
\hline
\end{tabular}
}
\hfill
\resizebox{0.5\textwidth}{!}{
\centering
\begin{tabular}{ | l | r | r | }
\hline
Method						&   Class Acc. 	\\
\hline
\hline
Vezhnevets et al.~\cite{vezhnevets11iccv}	&   	14.0\%	\\
Vezhnevets et al.~\cite{vezhnevets12cvpr}	&   	21.0\%	\\
Zhang et al.~\cite{zhang14tm}			&   	27.7\%	\\
Xu et al.~\cite{xu14cvpr}			&	27.9\%	\\
Zhang et al.~\cite{zhang15cvpr}			&	32.3\%	\\
Xu et al.~\cite{xu15cvpr}			&   	35.0\%	\\
\hline
Xu et al.~\cite{xu15cvpr}			&   	41.4\%	\\
(transductive)					&   		\\
						&   		\\
						&   		\\
						&   		\\
						&   		\\
\hline
Ours SVM \;\;\;\;\;\;(AlexNet)			&	21.2\% 	\\
Ours SVM+PS (AlexNet)				&	16.8\%	\\
Ours SVM+JC (AlexNet)				&	37.4\%  \\
Ours SVM+JC (VGG16)				& \textbf{44.8\%}\\
\hline 
\end{tabular}
}
\vspace{-3mm}
\caption{
Class-average pixel accuracy in the fully supervised (left) and the weakly supervised setting (right) setting.
We show results for our model on the test set of SIFT Flow using uncalibrated SVM scores (SVM), traditional Platt scaling~(PS) and joint calibration~(JC).
}
\label{tab:siftflow}
\end{table}

\mypar{Fully supervised setting.}
Table~\ref{tab:siftflow} evaluates various versions of our model in the fully supervised setting, and compares to other works on SIFT Flow.
Using AlexNet features and uncalibrated SVMs, our model achieves a class-average pixel accuracy of $28.7\%$.
If we calibrate the SVM scores with traditional Platt scaling results do not improve~($27.7\%$).
Using our proposed joint calibration to maximize class-average pixel accuracy improves results substantially to $55.6\%$.
This shows the importance of joint calibration to resolve conflicts between overlapping regions at multiple scales, to take into account competition between classes, and generally to optimize a loss mirroring the evaluation measure.

\begin{figure*}
\begin{center}
\includegraphics[width=0.95\textwidth]{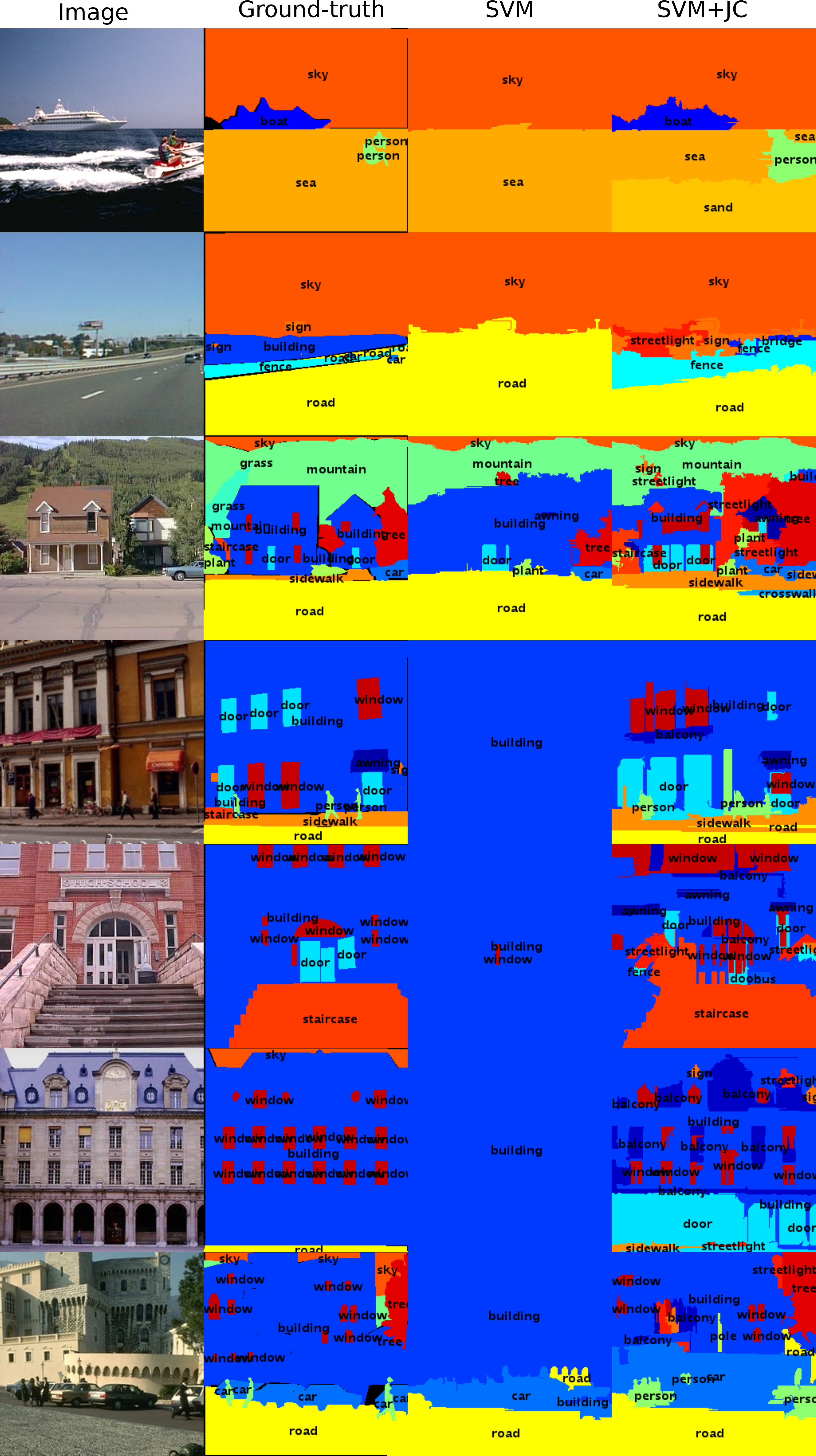}
\vspace{-6mm}
\end{center}
\caption{
   Fully supervised semantic segmentation on SIFT Flow.
   We present uncalibrated SVM results~(SVM) and jointly calibrated results~(SVM+JC),
   both with VGG16.
}
\label{fig:examples_sl}
\end{figure*}

Fig.~\ref{fig:examples_sl}~(column ``SVM'') shows that larger background regions~(i.e. sky, building) swallow smaller foreground regions~(i.e. boat, awning).
Many of these small objects become visible after calibration~(column ``SVM+JC'').
This issue is particularly evident when working with overlapping regions.
Consider a large region on a building which contains an awning.
As the surface of the awning is small, the features of the large region will be dominated by the building, leading to strong classification score for the `building' class.
When these are higher than the classification score for `awning' on the small awning region, the latter gets overwritten.
Instead, this problem does not appear when working with superpixels~\cite{boix12ijcv}.
A superpixel is either part of the building or part of the awning, so a high scoring awning superpixel cannot be overwritten by neighboring building superpixels.
Hence, joint calibration is particularly important when working with overlapping regions.

Using the deeper VGG16 CNN the results improve further, leading to our final performance $59.2\%$. This outperforms the state-of-the-art~\cite{sharma15cvpr} by $6.4\%$.

\mypar{Weakly supervised setting.}
Table~\ref{tab:siftflow} shows results in the weakly supervised setting.
The model with AlexNet and uncalibrated SVMs achieves an accuracy of 21.2\%.
Using traditional Platt scaling the result is 16.8\%, again showing it is not appropriate for semantic segmentation.
Instead, our joint calibration almost doubles accuracy (37.4\%).
Using the deeper VGG16 CNN results improve further to $44.8\%$.

\begin{figure*}
\begin{center}
\includegraphics[width=1\textwidth]{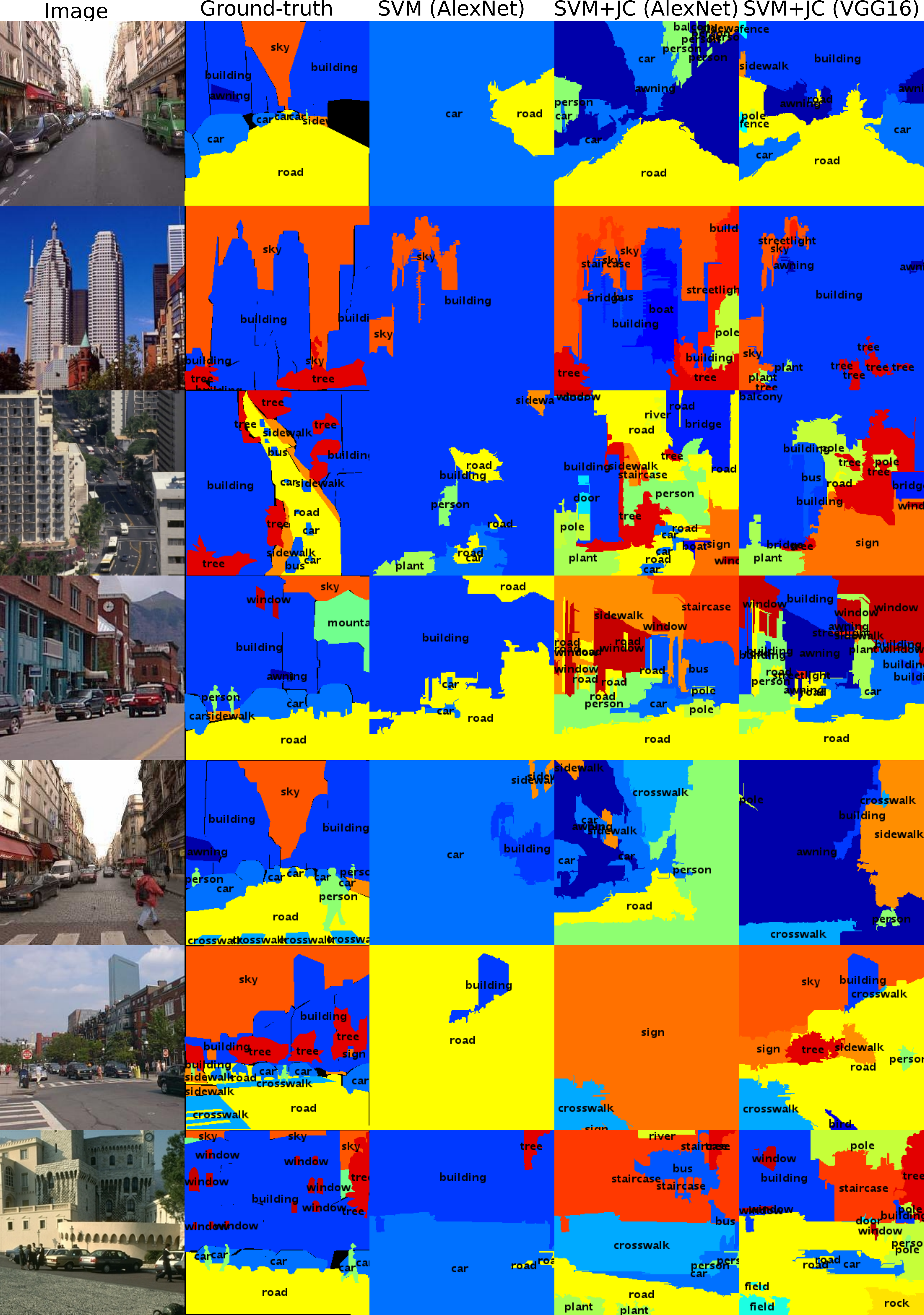}
\vspace{-6mm}
\end{center}
\caption{
   Weakly supervised semantic segmentation on SIFT Flow.
   We present uncalibrated SVM results~(SVM) with AlexNet, jointly calibrated results~(SVM+JC) with AlexNet, and with VGG16.
}
\vspace{-3mm}
\label{fig:examples_wsl}
\end{figure*}

Fig.~\ref{fig:examples_wsl} illustrates the power of our weakly supervised method.
Again rare classes appear only after joint calibration.
Our complete model outperforms the state-of-the-art~\cite{xu15cvpr} (35.0\%) in this setting by $9.8\%$.
Xu et al.~\cite{xu15cvpr} additionally report results on the transductive setting~($41.4\%$), where all (unlabeled) test images are given to the algorithm during training.

\mypar{Region proposals.}
To demonstrate the importance of multi-scale regions, we also analyze oversegmentations that do not
cover multiple scales. To this end, we keep our framework the same, but instead of Selective
Search~(SS)~\cite{uijlings13ijcv} region proposals we used a single oversegmentation using the 
method of Felzenszwalb and Huttenlocher~(FH)~\cite{felzenszwalb04ijcv} (for which we optimized the
scale parameter).
As Table~\ref{tab:regions} shows, SS regions outperform FH regions by a good margin of $12.2\%$ in the fully supervised setting.
This confirms that overlapping multi-scale regions are superior to non-overlapping oversegmentations.

\mypar{CNN finetuning.}
As described in~\ref{par:features} we finetune our network for detection in the fully supervised case.
Table~\ref{tab:finetuning} shows that this improves results by $6.2\%$ compared to using a CNN trained only for image classification on ILSVRC 2012.

\begin{table}[t]

\parbox{.45\linewidth}{
\centering
\begin{tabular}{ | l | r |}
\hline
Regions & Class Acc.	\\
\hline 
FH~\cite{felzenszwalb04ijcv}	& 	43.4\%	\\ 
SS~\cite{uijlings13ijcv}	& 	55.6\%	\\ 
\hline
\end{tabular}
\vspace{+1mm}
\caption{Comparison of single-scale~(FH) and multi-scale~(SS) regions using SVM+JC (AlexNet).}
\label{tab:regions}
}
\hfill
\parbox{.45\linewidth}{
\centering
\begin{tabular}{ | l | r |}
\hline
Finetuned	& 	Class Acc.	\\
\hline
no		&	49.4\%		\\
yes		&	55.6\%		\\
\hline
\end{tabular}
\vspace{+1mm}
\caption{Effect of CNN finetuning in the fully supervised setting using SVM+JC (AlexNet).}
\label{tab:finetuning}
}
\vspace{-2mm}
\end{table}

\vspace{-2mm}
\section{Conclusion}
\vspace{-2mm}

We addressed three common problems in semantic segmentation based on region proposals:
(1) overlapping regions yield conflicting class predictions at the pixel-level;
(2) class-imbalance leads to classifiers unable to detect rare classes;
(3) one-vs-all classifiers do not take into account competition between multiple classes.
We proposed a joint calibration strategy which optimizes a loss defined over the final pixel-level output labeling of the model, after maximization over classes and regions.
This tackles all three problems:
\emph{joint} calibration deals with multi-class predictions, while our loss explicitly deals with class imbalance and is defined in terms of pixel-wise labeling rather than region classification accuracy.
As a result we take into account conflict resolution between overlapping regions.
Our method outperforms the state-of-the-art in both the fully and the weakly supervised setting on the popular SIFT Flow~\cite{liu11pami} benchmark.

\mypar{Acknowledgements.} Work supported by the ERC Starting Grant VisCul.

\pagebreak
\bibliography{shortstrings,calvin,vggroup}

\end{document}